\documentclass{article}

\usepackage{PRIMEarxiv}

\usepackage[utf8]{inputenc} 
\usepackage[T1]{fontenc}    
\usepackage{hyperref}       
\usepackage{url}            
\usepackage{booktabs}       
\usepackage{amsfonts}       
\usepackage{nicefrac}       
\usepackage{microtype}      
\usepackage{lipsum}
\usepackage{booktabs}
\usepackage{fancyhdr}       
\usepackage{graphicx}       
\graphicspath{{media/}}     

\title{A Low-Cost Machine Learning Approach for Timber Diameter Estimation
}

\author{
   Fatemeh Hasanzadeh Fard \\
   Department of Wood and Paper Science and Technology \\
   University College of Agriculture and Natural Resources\\
   University of Tehran \\
   Tehran, Iran \\
   \texttt{f.hasanzadeh@ut.ac.ir}  
   \And
   Sanaz Hasanzadeh Fard \\
   Department of Computer Science and Engineering \\
   Michigan State University \\
   East Lansing, Michigan \\
   \texttt{hasanzad@msu.edu} 
   \And
   Mehdi Jonoobi \\
   Department of Wood and Paper Science and Technology \\
   University College of Agriculture and Natural Resources\\
   University of Tehran \\
   Tehran, Iran \\
   \texttt{mehdi.jonoobi@ut.ac.ir}
}

\begin{document}
\maketitle

\begin{abstract}
The wood processing industry, particularly in facilities such as sawmills and MDF production lines, requires accurate and efficient identification of species and thickness of the wood. Although traditional methods rely heavily on expert human labor, they are slow, inconsistent, and prone to error, especially when processing large volumes. This study focuses on practical and cost-effective machine learning frameworks that automate the estimation of timber log diameter using standard RGB images captured under real-world working conditions. We employ the YOLOv5 object detection algorithm, fine-tuned on a public dataset (TimberSeg 1.0), to detect individual timber logs and estimate thickness through bounding-box dimensions. Unlike previous methods that require expensive sensors or controlled environments, this model is trained on images taken in typical industrial sheds during timber delivery. Experimental results show that the model achieves a mean Average Precision (mAP@0.5) of 0.64, demonstrating reliable log detection even with modest computing resources. This lightweight, scalable solution holds promise for practical integration into existing workflows, including on-site inventory management and preliminary sorting, particularly in small and medium-sized operations.

\end{abstract}


\section{Introduction}
Timber is a key raw material in a wide range of industries, including construction, furniture manufacturing, and the production of engineered wood products \cite{svatovs2022advanced}. While all of these sectors rely on timber, the required type often varies by species and diameter \cite{liu2024review, hossain2012mechanical}. For instance, some species offer the structural strength needed for construction \cite{serrano2007fracture}, while others are better suited for paneling or paper production \cite{mbereyaho2019timber}. Accurately identifying wood characteristics is essential for maintaining consistent product quality and making efficient use of raw materials \cite{machado2019assessment}.


Traditionally, the task of timber classification \cite{de2020analysis} relies on human labor \cite{herzog2012timber}; species classification is done mainly by experts, while diameter categorization into valid ranges is performed by experienced workers using conventional tools. As mentioned earlier, the main issue with manual classification is that it is time-consuming. In addition, factors such as inconsistent lighting, similar surface features among species, and large volumes of material can further complicate the process. As production scales up, the need for more reliable and efficient solutions becomes increasingly crucial \cite{dormontt2015forensic}.


The emergence of artificial intelligence (AI) and machine learning (ML) has led to advancements across many domains \cite{fard2023machine}, from everyday use cases like real-time prediction in social networks, to face recognition and surveillance, and to efficient industrial solutions for resource management \cite{hasanzadeh2019two, zhao2003face, fard2023temporal}. Natural science domains are no exception; they include a wide range of tasks such as fruit sorting and agricultural product packaging \cite{dewi2020fruit}, wood protection \cite{poshtiri2024functionalized}, species classification, and biological, ecological, and evolutionary analysis \cite{fard2025robustness}.

Some industries benefited from machine learning early on, but not all domains adopted it at the same pace. Forestry and agriculture are among those slower to adapt. Advances in computer vision \cite{szeliski2022computer, ballard1982computer, stockman2001computer, voulodimos2018deep} have opened opportunities to automate aspects of timber handling and inspection \cite{apolinario2019open, kumar2024timber, ding2021sawn}. Image processing methods help automate timber-related tasks, reduce dependence on manual labor, and improve consistency in identifying key characteristics. However, building systems that are accurate, cost-effective, and adaptable to different working environments remains an ongoing challenge.

In this study, we explore how AI and ML can support forestry and environmental science through a simple, low-cost application. Specifically, we present a straightforward framework for timber diameter estimation, as a continuation of the author's previous work \cite{fard2023machine} on the broader use of ML in scientific and industrial contexts.
Our approach uses a standard machine learning model to detect timber logs and estimate their diameter from RGB images. The goal is not to introduce a new algorithm, but to show that even existing tools can be effectively applied in this domain with minimal resources. By combining object detection with basic size estimation, we offer a practical and adaptable solution that works without specialized hardware or complex setups.


The rest of this paper is organized as follows: Section 2 reviews previous research in the field. Section 3 describes the technique, the proposed framework, and the dataset used. Section 4 presents the experimental results, followed by a discussion in Section 5. Finally, Section 6 provides the conclusion and outlines possible directions for future research and improvements.

\section{Literature Review}
The use of machine learning and computer vision in forestry \cite{alif2024yolov1, liu2018application, zhao2019comparison, alkhatib2023brief, estrada2023machine} has gained increasing attention as the industry looks for ways to improve efficiency, reduce costs, and enhance the accuracy of wood classification and measurement \cite{estrada2023machine, nasir2019classification, van1984wood, vicari2019leaf, pratondo2022comparison, hu2019deep, de2020analysis}. Traditional timber sorting processes \cite{cass2009cost, murphy2015stand, kunickaya2022analysis, zyrjanov2021modeling, taube2020effect, kizha2016processing} have typically relied on manual inspection, which depends heavily on the expertise of human operators. While this approach can be effective, it is time-consuming and often inconsistent due to fatigue or subjective judgment. To address these challenges, researchers have explored automated systems that leverage imaging technologies and statistical models.

Earlier studies in this area often focused on texture and grain analysis to classify wood species. For example, microscopic imaging and hand-crafted feature extraction methods such as Local Binary Patterns (LBP) \cite{rosa2022improved, yadav2015multiresolution, maruyama2018automatic} or Gray Level Co-occurrence Matrices (GLCM) \cite{bremananth2009wood, wang2010wood} have been used to distinguish between species with high accuracy in controlled environments. Although these techniques perform well on clean, prepared samples, they tend to struggle with raw timber images captured in more realistic settings, where lighting and surface variations are common.

More recently, the introduction of convolutional neural networks (CNNs) \cite{li2021survey, o2015introduction} has shifted the focus toward end-to-end learning approaches. CNN-based models have shown strong performance in wood recognition tasks \cite{holmstrom2023tree, nguyen2023evaluation, kim2024performance}, especially when trained on large collections of labeled images. Some studies have used ResNet or VGG architectures \cite{ergun2024wood, sun2021wood} to classify cross-sectional wood images and achieved notable accuracy improvements compared to earlier feature-based methods. However, these models often rely on close-up images of prepared wood sections, which limits their use in more practical environments like log yards or sawmills, where logs are stacked, partially occluded, or affected by environmental conditions.

Alongside classification, object detection has become a key area of research for identifying and locating timber in complex scenes. The YOLO family of algorithms \cite{diwan2023object}, in particular, has been widely used for real-time detection tasks due to its speed and relatively straightforward implementation. YOLO models have also been applied in agriculture \cite{badgujar2024agricultural, alif2024yolov1} for fruit counting, plant disease detection, and livestock monitoring—demonstrating that real-time object detection can be effective in various natural settings. However, the use of YOLO for timber detection and measurement is still relatively underexplored, with most prior work focusing on log counting or relying on specialized equipment like laser scanners or 3D cameras.

A related area involves the estimation of geometric properties such as diameter or volume. Techniques based on 3D scanning, LiDAR, or structured light sensors can provide detailed measurements and have been integrated into some commercial timber processing systems \cite{alvites2022lidar, akay2009using}. For example, laser-based scanning rigs can generate precise 3D models of logs for grading and cutting optimization. While these systems deliver excellent accuracy, they require significant financial investment and are often impractical for small or medium-sized operations. By comparison, vision-based approaches using 2D images are far more affordable and easier to deploy, though they generally offer coarser measurements.

This work builds on that background by using a simple, low-cost method that leverages YOLOv5 \cite{jocher2020ultralytics} to detect timber logs in standard RGB images. Unlike methods that rely on specialized sensors or highly controlled imaging setups, this approach is designed to work under typical field conditions. By combining object detection with a basic width-based diameter estimation method, the system offers a practical compromise between measurement precision and accessibility. Although it cannot determine exact log diameters and only classifies them into broad ranges, it can serve as a useful preprocessing tool.

\section{Methodology}
\subsection{Overview}
Our framework focuses on the primary task of timber diameter estimation, based on object detection applied to RGB images of timber pieces. The pipeline is structured around a single model that detects and classifies individual timber objects in each image using a modified YOLOv5 architecture. Diameter is either treated as a separate classification task (e.g., thin, medium, thick) or inferred from bounding box dimensions.

\subsection{Data Collection and Annotation}
\begin{figure}
    \centering
    \includegraphics[width=0.5\linewidth]{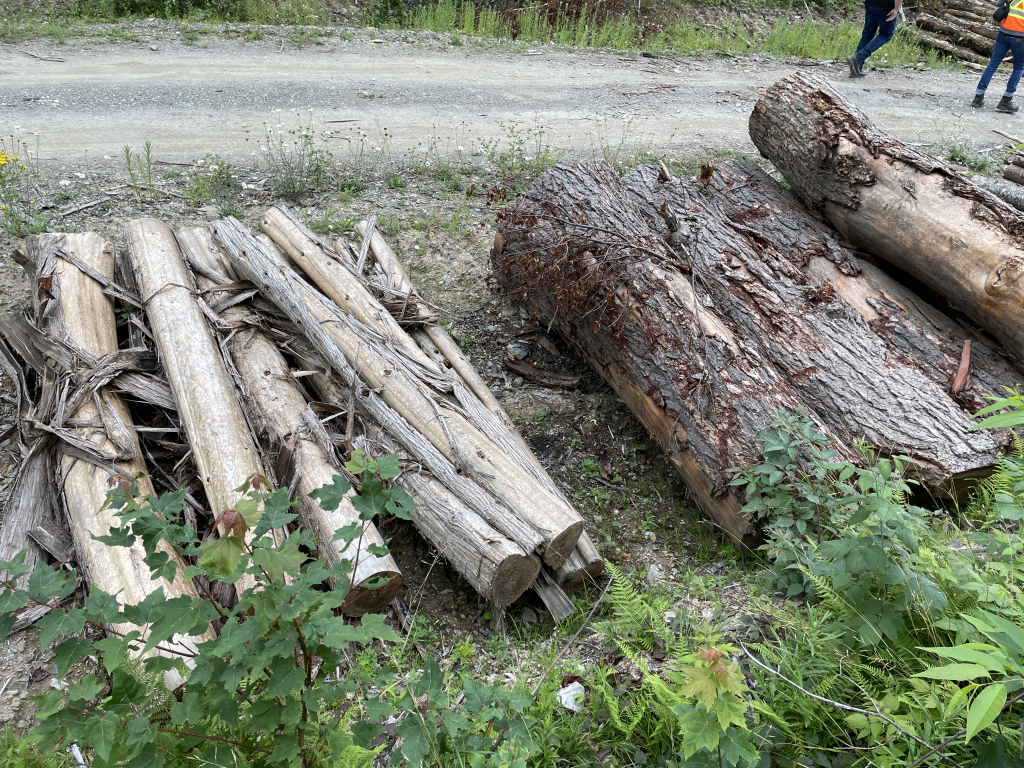}
    \caption{A sample image from the TimberSeg 1.0 dataset}
    \label{first}
\end{figure}

We used a set of timber images collected from the publicly available TimberSeg 1.0 dataset \cite{fortin2022instance} and performed annotations compatible with the YOLOv5 format using tools provided by \href{https://universe.roboflow.com/}{Roboflow Universe}, which includes wood recognition and annotated log detection sets. Images were selected to represent a range of diameters. Figure~\ref{first} shows a sample image from the TimberSeg 1.0 dataset. Table~\ref{first_t} presents TimberSeg 1.0 statistics.



\begin{table}[h!]
\centering
\normalsize
\renewcommand{\arraystretch}{1.2}
\caption{Overview of TimberSeg 1.0 Dataset Statistics}
\begin{tabular}{l@{\hskip 3em}c}
\textbf{Attribute} & \textbf{Value} \\
\midrule
Total Images & 440 (220 original + 220 prescaled) \\
Total Annotated Log Instances & 10,735 \\
Average Logs per Image & 24.4 \\
Number of Classes & 1 (Wood) \\
Annotation Types & Dense segmentation masks, bounding boxes \\
Average Object Area & ~2.85\% of image area \\
Object Area Range & 0\% to 78\% of image area \\
Object Width Range & 2 px to 4,608 px \\
Object Height Range & 2 px to 3,489 px \\
Image Sources & Dashcams (forest, roadside, trailer) \\
Purpose & Instance segmentation for log detection \\
\end{tabular}
\label{first_t}
\end{table}






Labels are as follows: a diameter category defined by visual estimate or bounding box width (e.g., Thin <30px, Medium 30–60px, Thick >60px). Annotations were performed using Roboflow and exported in YOLOv5 format for compatibility with the detection framework.

\subsection{Model Architecture}
We use a pre-trained YOLOv5 model as the base detector. The model was fine-tuned on our dataset using transfer learning \cite{torrey2010transfer, weiss2016survey}. Diameter estimation is handled as a post-processing step, using bounding box width to assign logs to diameter categories.

Training was carried out in Google Colab. A batch size of 8–16 and an image resolution of 416×416 were used. Models were trained for 50–100 epochs with early stopping enabled.

\begin{figure}
    \centering
    \includegraphics[width=0.9\linewidth]{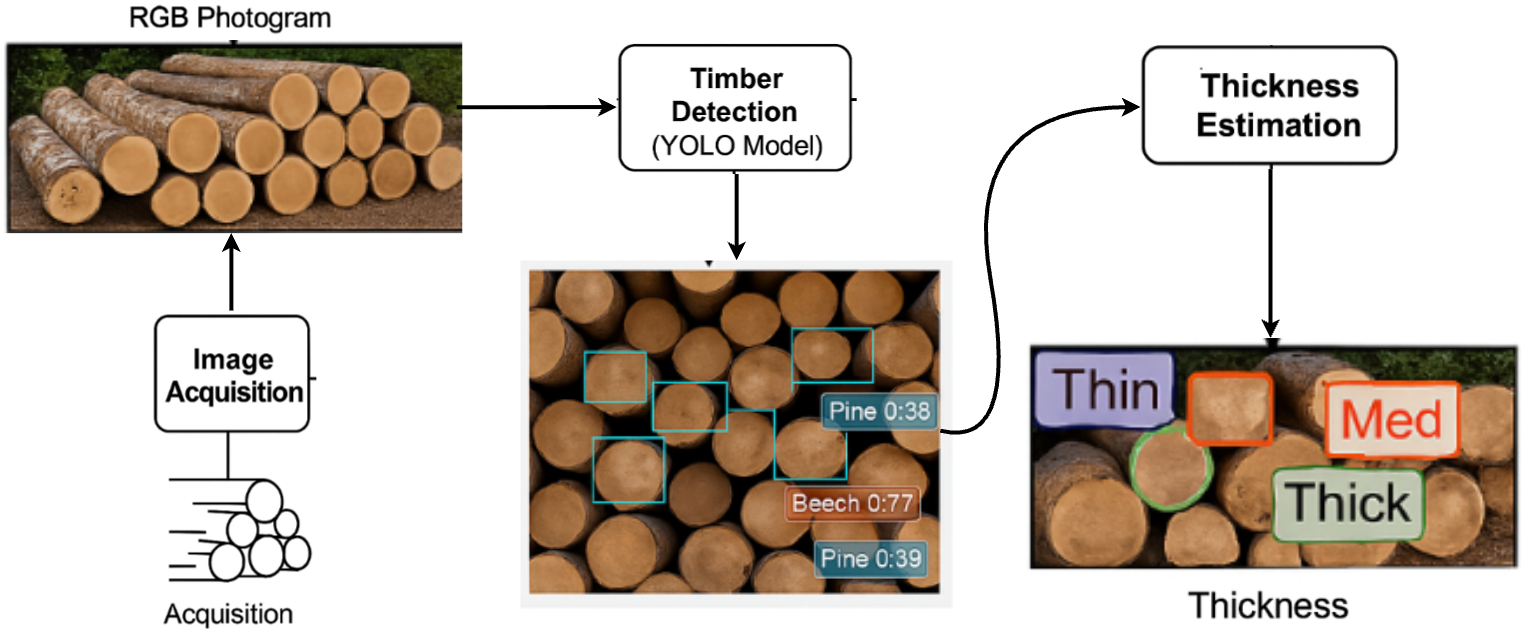}
    \caption{Overview of the proposed timber detection and diameter estimation workflow, using a YOLO-based model.}
    \label{three}
\end{figure}

Figure~\ref{three} illustrates a high-level view of the timber detection and diameter estimation framework.

\subsection{Evaluation Metrics}
Diameter estimation was assessed using bin accuracy (i.e., correct classification into the defined diameter ranges) and comparison with manually labeled values. Overall detection performance was measured using mAP@0.5 and IOU thresholds.

\section{Results}
The model was trained using YOLOv5 on a subset of timber log images from the TimberSeg 1.0 dataset, annotated for object detection. The training process was completed in 50 epochs on Google Colab using a CPU-only environment. Despite the modest setup, the model converged successfully, with training and validation losses decreasing steadily.

\begin{figure}
    \centering
    \includegraphics[width=1\linewidth]{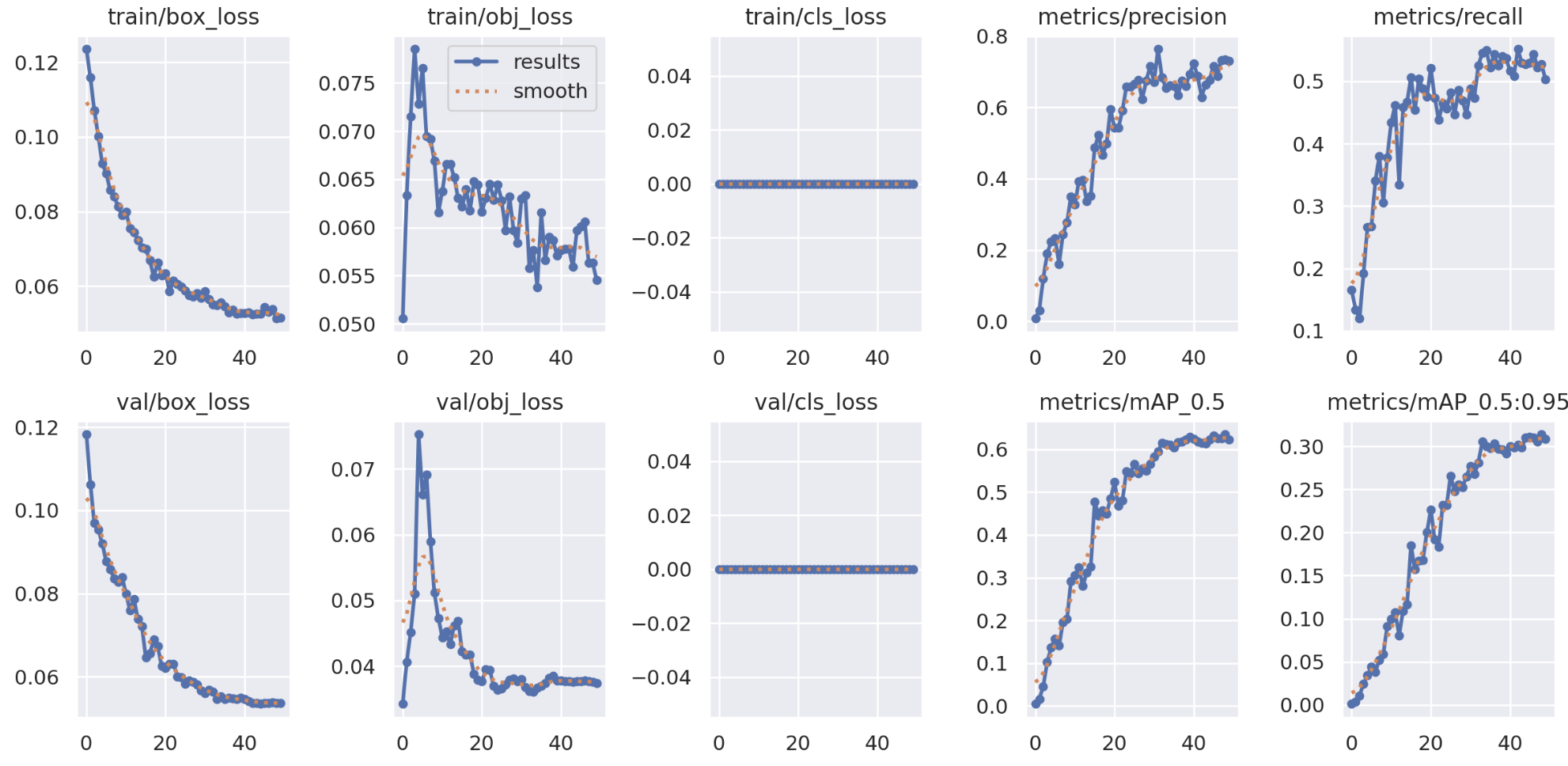}
    \caption{Experimental results}
    \label{second}
\end{figure}

On the validation set, the model achieved a precision of approximately 0.75 and a mean Average Precision (mAP@0.5) of 0.60. When tested on a held-out test set of 21 images containing 208 labeled log instances, the model maintained strong performance, with a precision of 0.656, recall of 0.577, mAP@0.5 of 0.640, and mAP@0.5:0.95 of 0.356. Figure~\ref{second} shows sample experimental results. These metrics indicate good generalization to unseen data, particularly considering the small dataset and the single-class detection setup.

Inference speed was also acceptable for real-time or near-real-time use cases, with average processing times under 300ms per image on CPU. This makes the system viable for applications where speed is important, such as on-site timber sorting or inventory scanning.

In addition to numerical metrics, qualitative results supported the model’s effectiveness. Sample output images from the test set showed accurate bounding boxes around individual logs, with minimal false positives or missed detections. The model was also able to assign estimated diameter ranges based on bounding box size, which appeared visually consistent across multiple test images.

While the model was trained to detect only one class (“log”), it handled varying orientations, partial occlusions, and background changes fairly well. Some degradation in accuracy was observed in cases of overlapping logs or poor lighting, which highlights the importance of dataset diversity.

\section{Discussion}

The trained model performed reasonably well, achieving solid results given the simplicity of the setup and the size of the dataset. The final mAP@0.5 on the test set was 0.640, indicating that the model could effectively detect and localize logs in new images. The decision to use YOLOv5 turned out to be practical—it allowed for fast training and inference without requiring a powerful GPU, making the system easier to deploy in field settings or small-scale operations.

However, there are several important aspects to reflect on. First, while the model was able to distinguish logs from the background accurately, it only considered one object class. In practice, wood processing often requires knowing not just where the logs are, but also their species and quality class. The current model does not yet support that level of differentiation.

\newpage

Another challenge lies in the method used for estimating diameter: since it relies on the width of bounding boxes in 2D images, results can be affected by camera angles, overlapping objects, or inconsistent scaling, which may reduce accuracy in real-world use.

In addition, the training dataset was relatively small and not very diverse. It came from a controlled source and lacked environmental variation such as shadows, mixed lighting, or background clutter. For broader deployment, the model would likely need retraining or fine-tuning on new images from the intended use environment.

Despite these limitations, the work shows promise. The pipeline works end-to-end, produces consistent results, and is flexible enough to be extended. Adding more object classes (e.g., species or quality), combining RGB with depth data, or integrating the system into an automated conveyor line are all potential next steps for improving both performance and practical value.

\section{Conclusion and Future Work}
This work presents an initial exploration into using computer vision and deep learning to assist with timber detection and categorization tasks. A YOLOv5-based object detection model was trained to detect logs and estimate their diameter range using standard RGB images, without the need for special sensors or manual measurements. The system is low-cost, relatively simple to implement, and capable of producing practical results in real time.

Evaluation on a test set showed that the model achieved a precision of 0.656 and an mAP@0.5 of 0.640—promising results considering the limited size of the dataset. These findings suggest that machine learning techniques can be a useful tool in automating parts of the wood processing pipeline, especially in settings where manual labor is time-consuming or inconsistent.

To better understand the future directions of this research, it is important to first acknowledge the model’s limitations.  
First, the model was trained exclusively on the TimberSeg 1.0 dataset, which may lack the environmental diversity found in real-world sawmill or yard conditions (e.g., variable lighting, occlusion, background clutter). This limits the system’s generalization; one direction for future work is to include images captured in operational environments with varied lighting and background complexity to improve robustness.

A second limitation is that diameter estimation is based solely on the bounding box width in 2D RGB images. This method is prone to distortion due to perspective, camera angle, and overlapping logs, all of which can reduce accuracy in real use. Incorporating depth estimation techniques (e.g., stereo vision, RGB-D sensors) or applying perspective correction could improve measurement accuracy.

\newpage

Another consideration is that the current detection model uses only one class (“log”), and diameter is either treated as a separate label or inferred post-hoc. This limits its usefulness in tasks requiring more detailed log profiling (e.g., species, size, or quality). Implementing multi-class detection or a multi-task model (e.g., species and diameter jointly) could increase its practical value.

It is also worth noting that macroscopic timber identification becomes more difficult in the presence of fungal decay, discoloration (e.g., false heartwood), or other physiological defects. These conditions can significantly alter the surface texture and color patterns, leading to possible misclassification by both human experts and machine learning models.

While there are still areas for improvement—such as enhancing the diameter estimation method, incorporating timber species into the model, and expanding the dataset—this work provides a functional base. It demonstrates that even with a modest dataset and computational setup, it is possible to train a model that offers valuable insights for industrial or forestry-related use cases.

In the future, this system could be adapted and extended for real-time classification, inventory management, or even automated quality control, helping to reduce waste and optimize material usage.




\bibliographystyle{unsrt}  
\bibliography{references}

\begin{thebibliography}{10}

\bibitem{svatovs2022advanced}
Hana Svato{\v{s}}-Ra{\v{z}}njevi{\'c}, Luis Orozco, and Achim Menges.
\newblock Advanced timber construction industry: A review of 350 multi-storey timber projects from 2000--2021.
\newblock {\em Buildings}, 12(4):404, 2022.

\bibitem{liu2024review}
Julie Liu and Erica~C Fischer.
\newblock Review of the charring rates of different timber species.
\newblock {\em Fire and Materials}, 48(1):3--15, 2024.

\bibitem{hossain2012mechanical}
M~Bellal Hossain and ASM Abdul Awal ASM~Abdul Awal.
\newblock Mechanical properties and durability of some selected timber species.
\newblock {\em Malaysian Journal of Civil Engineering}, 24(1), 2012.

\bibitem{serrano2007fracture}
Erik Serrano and Per~Johan Gustafsson.
\newblock Fracture mechanics in timber engineering--strength analyses of components and joints.
\newblock {\em Materials and structures}, 40:87--96, 2007.

\bibitem{mbereyaho2019timber}
Leopold Mbereyaho, Samuel Tuyishime, J~de~Montfort, TK~Uwintwali, and Chantal Tumukunde.
\newblock Timber application in construction industry and its promotion.
\newblock {\em Mediterranean Journal of Basic and Applied Sciences (MJBAS)}, 3(3):145--154, 2019.

\bibitem{machado2019assessment}
Jose~Saporiti Machado, Filipe Pereira, and Teresa Quilho.
\newblock Assessment of old timber members: Importance of wood species identification and direct tensile test information.
\newblock {\em Construction and Building Materials}, 207:651--660, 2019.

\bibitem{de2020analysis}
Andr{\'e}~R de~Geus, S{\'e}rgio F~da Silva, Alexandre~B Gontijo, Fl{\'a}vio~O Silva, Marcos~A Batista, and Jefferson~R Souza.
\newblock An analysis of timber sections and deep learning for wood species classification.
\newblock {\em Multimedia Tools and Applications}, 79(45):34513--34529, 2020.

\bibitem{herzog2012timber}
Thomas Herzog, Julius Natterer, Roland Schweitzer, Michael Volz, and Wolfgang Winter.
\newblock {\em Timber construction manual}.
\newblock Walter de Gruyter, 2012.

\bibitem{dormontt2015forensic}
Eleanor~E Dormontt, Markus Boner, Birgit Braun, Gerhard Breulmann, Bernd Degen, Edgard Espinoza, Shelley Gardner, Phil Guillery, John~C Hermanson, Gerald Koch, et~al.
\newblock Forensic timber identification: It's time to integrate disciplines to combat illegal logging.
\newblock {\em Biological Conservation}, 191:790--798, 2015.

\bibitem{fard2023machine}
Sanaz~Hasanzadeh Fard.
\newblock Machine learning on dynamic graphs: a survey on applications.
\newblock {\em 2023 IEEE Ninth Multimedia Big Data (BigMM)}, pages 32--39, 2023.

\bibitem{hasanzadeh2019two}
Sanaz Hasanzadeh~Fard and Hadi Tabatabaee~Malazi.
\newblock A two-dimensional self-coordination mechanism of agents in a minority game.
\newblock In {\em International Conference On Fundamentals Of Software Engineering}, pages 20--36. Springer, 2019.

\bibitem{zhao2003face}
Wenyi Zhao, Rama Chellappa, P~Jonathon Phillips, and Azriel Rosenfeld.
\newblock Face recognition: A literature survey.
\newblock {\em ACM computing surveys (CSUR)}, 35(4):399--458, 2003.

\bibitem{fard2023temporal}
Sanaz~Hasanzadeh Fard and Mohammad Ghassemi.
\newblock Temporal link prediction using graph embedding dynamics.
\newblock In {\em 2023 IEEE Ninth Multimedia Big Data (BigMM)}, pages 48--55. IEEE, 2023.

\bibitem{dewi2020fruit}
Tresna Dewi, Pola Risma, and Yurni Oktarina.
\newblock Fruit sorting robot based on color and size for an agricultural product packaging system.
\newblock {\em Bulletin of Electrical Engineering and Informatics}, 9(4):1438--1445, 2020.

\bibitem{poshtiri2024functionalized}
Azar~Haghighi Poshtiri, Sima Sepahvand, Mehdi Jonoobi, Alireza Ashori, Ali~Naghi Karimi, Fatemeh~Hasanzadeh Fard, Laura Bergamonti, Claudia Graiff, and Sabrina Palanti.
\newblock Functionalized cellulose nanocrystals for enhanced wood protection: Synthesis, characterization, and performance.
\newblock {\em Industrial Crops and Products}, 222:120021, 2024.

\bibitem{fard2025robustness}
Sanaz~Hasanzadeh Fard and Emily Dolson.
\newblock The robustness of structural features in species interaction networks.
\newblock {\em arXiv preprint arXiv:2502.16778}, 2025.

\bibitem{szeliski2022computer}
Richard Szeliski.
\newblock {\em Computer vision: algorithms and applications}.
\newblock Springer Nature, 2022.

\bibitem{ballard1982computer}
Dana~Harry Ballard and Christopher~M Brown.
\newblock {\em Computer vision}.
\newblock Prentice Hall Professional Technical Reference, 1982.

\bibitem{stockman2001computer}
George Stockman and Linda~G Shapiro.
\newblock {\em Computer vision}.
\newblock Prentice Hall PTR, 2001.

\bibitem{voulodimos2018deep}
Athanasios Voulodimos, Nikolaos Doulamis, Anastasios Doulamis, and Eftychios Protopapadakis.
\newblock Deep learning for computer vision: A brief review.
\newblock {\em Computational intelligence and neuroscience}, 2018(1):7068349, 2018.

\bibitem{apolinario2019open}
Marco Paul~E Apolinario, Daniel A~Urcia Paredes, and Samuel G~Huaman Bustamante.
\newblock Open set recognition of timber species using deep learning for embedded systems.
\newblock {\em IEEE Latin America Transactions}, 17(12):2005--2012, 2019.

\bibitem{kumar2024timber}
P~Kumar, Sathish~Kumar Kannaiah, et~al.
\newblock Timber identification based on the grain structure using learning algorithms.
\newblock In {\em 2024 International Conference on Smart Technologies for Sustainable Development Goals (ICSTSDG)}, pages 1--6. IEEE, 2024.

\bibitem{ding2021sawn}
Fenglong Ding, Ying Liu, Zilong Zhuang, and Zhengguang Wang.
\newblock A sawn timber tree species recognition method based on am-sppresnet.
\newblock {\em Sensors}, 21(11):3699, 2021.

\bibitem{alif2024yolov1}
Mujadded Al~Rabbani Alif and Muhammad Hussain.
\newblock Yolov1 to yolov10: A comprehensive review of yolo variants and their application in the agricultural domain.
\newblock {\em arXiv preprint arXiv:2406.10139}, 2024.

\bibitem{liu2018application}
Zelin Liu, Changhui Peng, Timothy Work, Jean-Noel Candau, Annie DesRochers, and Daniel Kneeshaw.
\newblock Application of machine-learning methods in forest ecology: recent progress and future challenges.
\newblock {\em Environmental Reviews}, 26(4):339--350, 2018.

\bibitem{zhao2019comparison}
Qingxia Zhao, Shichuan Yu, Fei Zhao, Linghong Tian, and Zhong Zhao.
\newblock Comparison of machine learning algorithms for forest parameter estimations and application for forest quality assessments.
\newblock {\em Forest Ecology and Management}, 434:224--234, 2019.

\bibitem{alkhatib2023brief}
Ramez Alkhatib, Wahib Sahwan, Anas Alkhatieb, and Brigitta Sch{\"u}tt.
\newblock A brief review of machine learning algorithms in forest fires science.
\newblock {\em Applied Sciences}, 13(14):8275, 2023.

\bibitem{estrada2023machine}
Juan~Sebasti{\'a}n Estrada, Andr{\'e}s Fuentes, Pedro Reszka, and Fernando Auat~Cheein.
\newblock Machine learning assisted remote forestry health assessment: a comprehensive state of the art review.
\newblock {\em Frontiers in plant science}, 14:1139232, 2023.

\bibitem{nasir2019classification}
Vahid Nasir, Sepideh Nourian, Stavros Avramidis, and Julie Cool.
\newblock Classification of thermally treated wood using machine learning techniques.
\newblock {\em Wood Science and Technology}, 53:275--288, 2019.

\bibitem{van1984wood}
Ger~JCM Van~Vliet and Pieter Baas.
\newblock Wood anatomy and classification of the myrtales.
\newblock {\em Annals of the Missouri Botanical Garden}, pages 783--800, 1984.

\bibitem{vicari2019leaf}
Matheus~B Vicari, Mathias Disney, Phil Wilkes, Andrew Burt, Kim Calders, and William Woodgate.
\newblock Leaf and wood classification framework for terrestrial lidar point clouds.
\newblock {\em Methods in Ecology and Evolution}, 10(5):680--694, 2019.

\bibitem{pratondo2022comparison}
Agus Pratondo and Astri Novianty.
\newblock Comparison of wood classification using machine learning.
\newblock In {\em 2022 IEEE 10th Conference on Systems, Process \& Control (ICSPC)}, pages 308--312. IEEE, 2022.

\bibitem{hu2019deep}
Junfeng Hu, Wenlong Song, Wei Zhang, Yafeng Zhao, and Alper Yilmaz.
\newblock Deep learning for use in lumber classification tasks.
\newblock {\em Wood Science and Technology}, 53:505--517, 2019.

\bibitem{cass2009cost}
Randy~D Cass, Shawn~A Baker, and W~Dale Greene.
\newblock Cost and productivity impacts of product sorting on conventional ground-based timber harvesting operations.
\newblock {\em Forest Products Journal}, 59(11-12):108--114, 2009.

\bibitem{murphy2015stand}
Glen Murphy and Dave Cown.
\newblock Stand, stem and log segregation based on wood properties: a review.
\newblock {\em Scandinavian Journal of Forest Research}, 30(8):757--770, 2015.

\bibitem{kunickaya2022analysis}
OA~Kunickaya, Aleksandr Pomiguev, IN~Kruchinin, Tamara Storodubtseva, AM~Voronova, Dmitry Levushkin, Vyacheslav Borisov, and Viktor Ivanov.
\newblock Analysis of modern wood processing techniques in timber terminals.
\newblock {\em Central European Forestry Journal}, 68(1):51--59, 2022.

\bibitem{zyrjanov2021modeling}
Mihail Zyrjanov, Sergey Medvedev, and Tatiana Rjabova.
\newblock Modeling of the process of collection, sorting and transportation of logging residues at the logging area.
\newblock {\em Journal of Applied Engineering Science}, 19(1):114--118, 2021.

\bibitem{taube2020effect}
Piotr Taube, Kazimierz Or{\l}owski, Daniel Chucha{\l}a, and Jakub Sandak.
\newblock The effect of log sorting strategy on the forecasted lumber value after sawing pine wood.
\newblock {\em Acta Facultatis Xylologiae Zvolen}, 62:89--102, 2020.

\bibitem{kizha2016processing}
Anil~Raj Kizha and Han-Sup Han.
\newblock Processing and sorting forest residues: Cost, productivity and managerial impacts.
\newblock {\em Biomass and Bioenergy}, 93:97--106, 2016.

\bibitem{rosa2022improved}
N{\'u}bia Rosa~da Silva, Victor Deklerck, Jan~M Baetens, Jan Van~den Bulcke, Maaike De~Ridder, M{\'e}lissa Rousseau, Odemir~Martinez Bruno, Hans Beeckman, Joris Van~Acker, Bernard De~Baets, et~al.
\newblock Improved wood species identification based on multi-view imagery of the three anatomical planes.
\newblock {\em Plant Methods}, 18(1):79, 2022.

\bibitem{yadav2015multiresolution}
Arvind~R Yadav, Radhey~Shyam Anand, ML~Dewal, and Sangeeta Gupta.
\newblock Multiresolution local binary pattern variants based texture feature extraction techniques for efficient classification of microscopic images of hardwood species.
\newblock {\em Applied Soft Computing}, 32:101--112, 2015.

\bibitem{maruyama2018automatic}
Teruo~M Maruyama, LS~Oliveira, AS~Britto~Jr, and Silvana Nisgoski.
\newblock Automatic classification of native wood charcoal.
\newblock {\em Ecological Informatics}, 46:1--7, 2018.

\bibitem{bremananth2009wood}
R~Bremananth, B~Nithya, and R~Saipriya.
\newblock Wood species recognition using glcm and correlation.
\newblock In {\em 2009 International Conference on Advances in Recent Technologies in Communication and Computing}, pages 615--619. IEEE, 2009.

\bibitem{wang2010wood}
Bi-hui Wang, Hang-jun Wang, and Heng-nian Qi.
\newblock Wood recognition based on grey-level co-occurrence matrix.
\newblock In {\em 2010 International Conference on Computer Application and System Modeling (ICCASM 2010)}, volume~1, pages V1--269. IEEE, 2010.

\bibitem{li2021survey}
Zewen Li, Fan Liu, Wenjie Yang, Shouheng Peng, and Jun Zhou.
\newblock A survey of convolutional neural networks: analysis, applications, and prospects.
\newblock {\em IEEE transactions on neural networks and learning systems}, 33(12):6999--7019, 2021.

\bibitem{o2015introduction}
Keiron O'shea and Ryan Nash.
\newblock An introduction to convolutional neural networks.
\newblock {\em arXiv preprint arXiv:1511.08458}, 2015.

\bibitem{holmstrom2023tree}
Eero Holmstr{\"o}m, Antti Raatevaara, Jonne Pohjankukka, Heikki Korpunen, and Jori Uusitalo.
\newblock Tree log identification using convolutional neural networks.
\newblock {\em Smart Agricultural Technology}, 4:100201, 2023.

\bibitem{nguyen2023evaluation}
Khanh Nguyen-Trong.
\newblock Evaluation of wood species identification using cnn-based networks at different magnification levels.
\newblock {\em International Journal of Advanced Computer Science and Applications}, 14(4), 2023.

\bibitem{kim2024performance}
Jong-Ho Kim, Wan-Geun Park, and Nam-Hun Kim.
\newblock Performance of convolutional neural network (cnn) and performance influencing factors for wood species classification of lepidobalanus growing in korea.
\newblock {\em Scientific Reports}, 14(1):18141, 2024.

\bibitem{ergun2024wood}
Halime Ergun.
\newblock Wood identification based on macroscopic images using deep and transfer learning approaches.
\newblock {\em PeerJ}, 12:e17021, 2024.

\bibitem{sun2021wood}
Yongke Sun, Qizhao Lin, Xin He, Youjie Zhao, Fei Dai, Jian Qiu, and Yong Cao.
\newblock Wood species recognition with small data: A deep learning approach.
\newblock {\em International Journal of Computational Intelligence Systems}, 14(1):1451--1460, 2021.

\bibitem{diwan2023object}
Tausif Diwan, G~Anirudh, and Jitendra~V Tembhurne.
\newblock Object detection using yolo: Challenges, architectural successors, datasets and applications.
\newblock {\em multimedia Tools and Applications}, 82(6):9243--9275, 2023.

\bibitem{badgujar2024agricultural}
Chetan~M Badgujar, Alwin Poulose, and Hao Gan.
\newblock Agricultural object detection with you only look once (yolo) algorithm: A bibliometric and systematic literature review.
\newblock {\em Computers and Electronics in Agriculture}, 223:109090, 2024.

\bibitem{alvites2022lidar}
Cesar Alvites, Marco Marchetti, Bruno Lasserre, and Giovanni Santopuoli.
\newblock Lidar as a tool for assessing timber assortments: A systematic literature review.
\newblock {\em Remote Sensing}, 14(18):4466, 2022.

\bibitem{akay2009using}
Abdullah~Emin Akay, Hakan O{\u{g}}uz, Ismail~Rakip Karas, and Kazuhiro Aruga.
\newblock Using lidar technology in forestry activities.
\newblock {\em Environmental monitoring and assessment}, 151:117--125, 2009.

\bibitem{jocher2020ultralytics}
Glenn Jocher, Alex Stoken, Jirka Borovec, Liu Changyu, Adam Hogan, Laurentiu Diaconu, Jake Poznanski, Lijun Yu, Prashant Rai, Russ Ferriday, et~al.
\newblock ultralytics/yolov5: v3. 0.
\newblock {\em Zenodo}, 2020.

\bibitem{fortin2022instance}
Jean-Michel Fortin, Olivier Gamache, Vincent Grondin, Fran{\c{c}}ois Pomerleau, and Philippe Giguere.
\newblock Instance segmentation for autonomous log grasping in forestry operations.
\newblock In {\em 2022 IEEE/RSJ International Conference on Intelligent Robots and Systems (IROS)}, pages 6064--6071. IEEE, 2022.

\bibitem{torrey2010transfer}
Lisa Torrey and Jude Shavlik.
\newblock Transfer learning.
\newblock In {\em Handbook of research on machine learning applications and trends: algorithms, methods, and techniques}, pages 242--264. IGI global, 2010.

\bibitem{weiss2016survey}
Karl Weiss, Taghi~M Khoshgoftaar, and DingDing Wang.
\newblock A survey of transfer learning.
\newblock {\em Journal of Big data}, 3:1--40, 2016.

\end{thebibliography}

\end{document}